\documentclass[runningheads,a4paper]{llncs}

\usepackage[ruled]{algorithm2e}
\usepackage{amssymb}
\usepackage{amsmath}
\usepackage{graphicx}
\usepackage{booktabs}

\usepackage{url}
\urldef{\mailsa}\path|{zhang.yi, wangzhengfei, xgx, |
\urldef{\mailsb}\path|lwx}@pku.edu.cn| 
\urldef{\mailsc}\path|{13910093298}@163.com| 
\newcommand{\keywords}[1]{\par\addvspace\baselineskip
\noindent\keywordname\enspace\ignorespaces#1}

\begin{document}

\mainmatter  % start of an individual contribution

\title{N2RPP: An Adversarial Network to Rebuild Plantar Pressure for ACLD Patients}

\titlerunning{N2RPP: An Adversarial Network to Rebuild Plantar Pressure}


\author{Yi Zhang$^{1}$
\footnotemark[1]
\and Zhengfei Wang$^{1}$\footnote[1]{Both authors contributed equally to this work.} \and Guoxiong Xu$^{1}$ \and Hongshi Huang$^{2}$ \and\\
Wenxin Li$^{1}$}

\authorrunning{N2RPP: An Adversarial Network to Rebuild Plantar Pressure}

\institute{School of Electronic Engineering and Computer Science, Peking University\\
No.5 Yiheyuan Road Haidian District, Beijing, P.R.China\\
\mailsa\\
\mailsb\\
\mailsc\\
\url{http://ai.pku.edu.cn}}


\toctitle{Lecture Notes in Computer Science}
\tocauthor{Authors' Instructions}
\maketitle

\begin{abstract}
Foot is a vital part of human, and lots of valuable information is embedded. Plantar pressure is one of which contains this information and it describes human walking features. It is proved that once one has trouble with lower limb, the distribution of plantar pressure will change to some degree. Plantar pressure can be converted into images according to some simple standards. In this paper, we take full advantage of these plantar pressure images for medical usage. We present N2RPP, a generative adversarial network (GAN) based method to rebuild plantar pressure images of anterior cruciate ligament deficiency (ACLD) patients from low dimension features, which are extracted from an autoencoder. Through the result of experiments, the extracted features are a useful representation to describe and rebuild plantar pressure images. According to N2RPP's results, we find out that there are several noteworthy differences between normal people and patients. This can provide doctors a rough direction of adjusting plantar pressure to a better distribution to reduce patients' sore and pain during the rehabilitation treatment for ACLD.
\end{abstract}

\section{Introduction}

Plantar pressure refers to pressure fields acting between the plantar surface of the foot and a supporting surface. It is employed in a wide range of applications including sports biomechanics and gait biometrics. In clinical research and application, much work concentrate on the foot and lower limb disease's impact on plantar pressure. Prior studies have proved that plantar pressure can guide to diagnose diabetic foot \cite{Caselli2002The}, anterior cruciate ligament injury \cite{Huang2015Anterior} and Parkinson disease \cite{Pihet2006}. These researches show that patients with disease above may change their walking pattern to some degree to avoid sore and pain, which leads to the distribution alteration of plantar pressure.

Several approaches in \cite{Stolwijk2010Plantar}, \cite{Keijsers2009A}, \cite{Lai2014Impact}, \cite{Cristina2011Spatiotemporal} and \cite{Cavanagh1987The} try to find the characteristics as much as possible to analyze plantar pressure data. However, these features cannot describe the whole plantar pressure without any loss. Therefore, it is vital to find out how to describe plantar pressure data more completely, which can help computer understand and analyze better and more efficient.

Another problem about plantar pressure is that once something abnormal occurred in patients' lower limb, they shall suffer sore and pain around their foot. If we can figure out the specific part of the foot and use some physical devices to help to simulate normal distribution, they may suffer less during the recovering process. A typical plantar pressure image with details is shown in Fig.1 in Supplement Material.

Our target disease this time is anterior cruciate ligament deficiency (ACLD), a debilitating sports injury that leads to altered knee loading, affecting performance of daily living activities and increasing the risk of early osteoarthritis of the knee \cite{Berchuck1990Gait} \cite{Gardinier2013Altered}, which has been proved to affect plantar pressure \cite{Huang2015Anterior}. 

The novelty of our work lays in the following aspects. First, we use autoencoder for plantar pressure images to extract representations which considered to describe the original images in minimum information loss. Second, using extracted representations to rebuild plantar pressure images and find out that there are several noteworthy differences between normal people and patients.

The rest of this paper is organized as follows. In section 2, we introduce our autoencoder architecture and generative adversarial models that we propose in this paper. In section 3 and 4, we explain our experiments using models introduced in section 2 and show the results in both statistical data and visualization. In section 5, we give a conclusion of the experiments’ results and discuss the possible direction for further work in this area.

\section{Methodology}

\subsection{Autoencoder}

In this section, we present an autoencoder for extracting features from original plantar pressure images. The features extraction's quality is mainly judged by the mean square error (MSE) between pixel-level original images and reconstructed images, and fine-tune using visual evaluation as an aid.

As is explained in the section 1, based on previous clinical research, doctors only focus on several points of plantar pressure data. However, these discrete man-made features lack the ability to unite as a whole for reconstruct a complete plantar pressure image.

\begin{figure}[htb]
\centering
\includegraphics[width=1.0\textwidth]{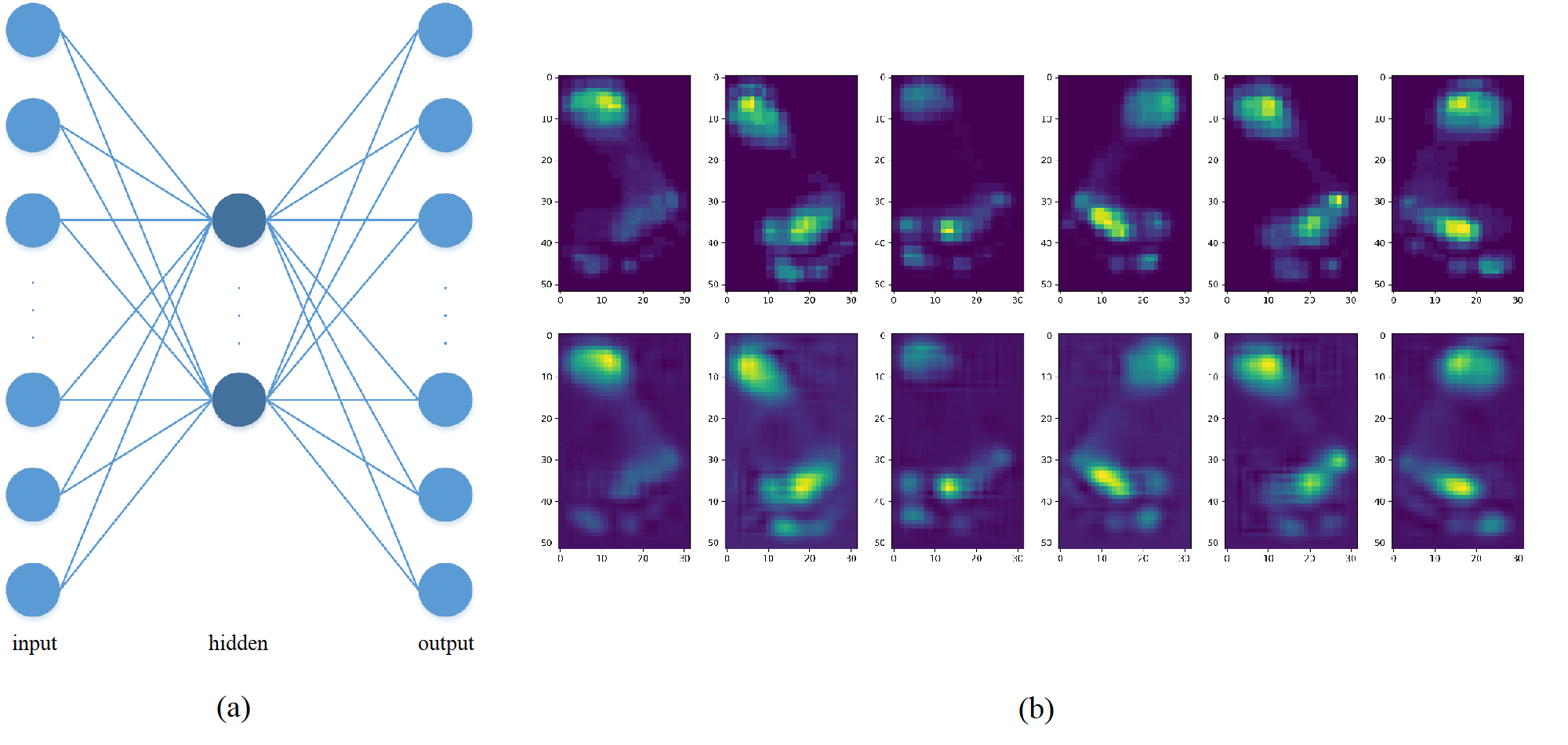}
\caption{(a) Autoencoder architecture. The input and output’s dimension is $1664(52\times 32)$, the intermediate vector’s dimension is 128. We extract the middle results as features. (b) Visual comparison between original plantar pressure images and outputs of autoencoder. The upper row is the original version and the below row is the reconstructive results.}
\label{fig:autoencoder}
\end{figure}

In order to obtain the features with least information loss, we draw inspiration from the breakthrough of deep learning and powerful ability of backpropagation. With some experiments, we finally deploy a one-hidden-layer symmetrical autoencoder framework for this features extraction task. Further, we fine-tune the autoencoder with visual evaluations and confirm the final configuration of the autoencoder. As is depicted in Fig.~\ref{fig:autoencoder}(a), the input and output dimension are 1664 and the features’ dimension is 128, generated by a fully-connected layer with rectified linear unit (ReLU) as activation function. The examples of input and output’s comparison are shown in Fig.~\ref{fig:autoencoder}(b).

\subsection{An Adversarial Network to Rebuild Plantar Pressure (N2RPP)}
\begin{figure}[tb]
\centering
\includegraphics[width=1.0\textwidth]{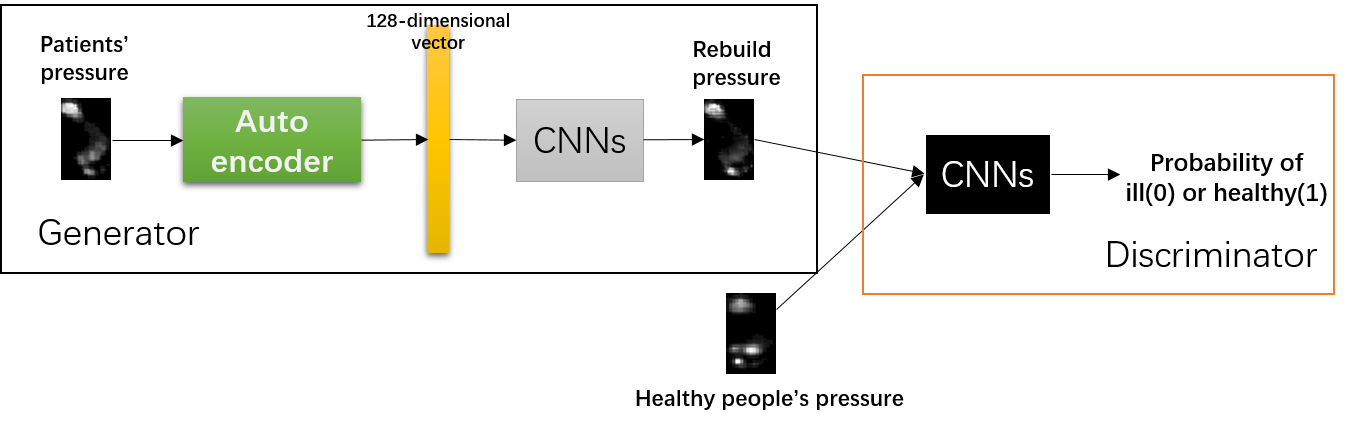}
\caption{The adversarial training procedure of N2RPP.}
\label{fig:procedure}
\end{figure}

%少一个图一个表
We adopt the general architecture of the deep convolutional generative adversarial networks (DCGANs) to rebuild the plantar pressure of ACLD patients. As a class of CNNs, DCGANs have certain architectural constraints. It demonstrated that DCGANs with some mild tweaking have more stable convergence compared to other GANs architectures\cite{radford2015unsupervised}. The procedure and architecture of N2RPP are shown in Fig.~\ref{fig:procedure}. A generator network is utilized to excavate the distribution rules within normal plantar pressure, while discriminator network distinguishes between the ground truth and the reconstructed image from the generator. For simplicity, we represent the generator and discriminator in N2RPP as $G$ and $D$, respectively. The detailed structures of $G$ and $D$ are shown in Tab.1 in Supplement Material. 

Our goal is to use the adversarial network to reconstruct the normal plantar pressure distribution for the patients, and assume that this distribution retains the inherent properties of the patients’ feet, walking gestures and etc. In order to make the reconstruction pressure as close as possible to the original plantar pressure distribution, we constrain an additional loss for the adversarial network. On the basis of the loss function of original GANs, we add a mean square error loss to measure the degree of difference. The new objective function is shown in the Eq.~\ref{eq:objective function} , in which $\alpha$ is used to control the weight of the two loss items. Optimizing the loss function of the adversarial model, we update the parameters of $G$ and $D$ from the backpropagated gradient.
\begin{equation}
\label{eq:objective function}
\begin{split}
  \min_{G}\max_{D} L(G, D) =& (1-\alpha)\left[\mathbb{E}_{y\sim p_{gt}}\log {D(y;\theta_d)} +
  \mathbb{E}_{x\sim p_{x}(x)}\log{(1-D(G(x;\theta_g);\theta_d))}\right] \\
  &+ \alpha \left[\mathbb{E}_{x\sim p_{x}(x)}(x-G(x;\theta_g))^2 \right]
\end{split}
\end{equation}

$G$ takes the patients’ original plantar pressure images as input and output the corrected pressure distribution. Input and output remain the same size. Next, we apply the autoencoder mentioned above to extract the features of the original input and obtain a 128-dimensional vector. The sparse features preserve some topological structures of the original pressure distribution, thus avoiding excessive redundant information, which may result in network collapse and fail to generate realistic images. The extracted vector is normalized to the Gaussian distribution with $\mu=0, \sigma=1$, so that it is more consistent with the random noise distribution. The normalized 128-dimensional vector is inputted into the deep convolution networks, and finally the reconstructed plantar pressure is obtained. It should be noted that the autoencoder is not trainable while training $G$ and $D$.

The loss function of training $G$ is shown as the Eq.~\ref{eq:g loss}. Minimizing the binary cross-entropy loss of the first component makes the output of $G$ more confusing to $D$. Minimization of the second component indicates the topology changes as small as possible. When training $G(x; \theta_g)$, $D$ is set to be non-trainable, and the label of $y_{rp}$ is assigned as one in order to make $D$ to mistake fake images as real. We replace the $\log(1-D(G(x)))$ in Eq.~\ref{eq:g loss} with $\log(D(G(x)))$ so as to provides much stronger gradients early in learning \cite{goodfellow2014generative}.
\begin{equation}
\label{eq:g loss}
\begin{split}
  l_G =& (1-\alpha)\left[\mathbb{E}_{x\sim p_{x}(x)}\log{(1-D(G(x;\theta_g);\theta_d))}\right] 
  + \alpha \left[\mathbb{E}_{x\sim p_{x}(x)}(x-G(x;\theta_g))^2 \right] \\
  =& (1-\alpha)\left[\mathbb{E}_{x\sim p_{x}(x)}\log{D(G(x;\theta_g);\theta_d)}\right] 
  + \alpha \left[\mathbb{E}_{x\sim p_{x}(x)}(x-G(x;\theta_g))^2 \right]
\end{split}
\end{equation}

For training $D(y;\theta_d)$, the ground truth label $y_{gt}$ is assigned as one, and the label of rebuild pressure $y_{rp} = G(x; \theta_g)$ is assigned as
zero where $x$ is the normalized vector of extracted features. Ground truth comes from the plantar pressure images of volunteers, and we guarantee that they do not have any history of disease. Training $D$ constrains by the loss function as below:
\begin{equation}
\label{eq:d loss}
  l_D = \mathbb{E}_{y\sim p_{gt}}\log {D(y;\theta_d)} +
  \mathbb{E}_{x\sim p_{x}(x)}\log{(1-D(G(x;\theta_g);\theta_d))}
\end{equation}

In accordance with the general idea of GANs, the purpose of training can be regarded as a confrontation game: $D$ learns to determine whether a sample is from the spurious distribution or the realistic distribution, while $G$ tries to produce fake data to cheat $D$ as far as possible. Once the accuracy of $D$ achieves around 50\%, the distribution of $G$ has almost recovered the ground truth distribution. Algorithm 1 in Supplement Material provides the scheme procedure for training the adversarial nets.

\section{Experiments}
% 少一个表
Unlike CT scan, emphysema, glaucoma and other diseases, anterior cruciate ligament deficiency (ACLD) does not have a public dataset. The data we use was captured from a plantar pressure acquisition system applied in cooperating hospital. Participants walk barefoot on a 0.5-meter Footscansystem (RSscan International, $0.48\times 0.32$m, 4,096 sensors and 2.6 sensors/$cm^2$)\footnote{\url{https://www.rsscan.com/footscan/entry-level/#tab_products}{}}, placed midway along a 10-meter walkway. The plantar pressure surface automatically generates a footprint rectangular module based on the landing site, which covers the foot position on the pressure surface. Walking a single trip we get a multi-frame pressure data containing two feet samples, each of which gives the pressure value of the effective point. Our data, collected from 2012 to 2017, includes patients of various ages, genders and occupations. Each patient is provided with a thorough diagnosis and medical history by sports medicine specialists.

% 这一段[1]需要替换成ICAPR的引用
After screening of the medical records, we select 1004 volunteer cases who have no medical history of sports functional impairment (each participant may have multiple cases, similarly hereinafter). In order to eliminate the cross-impact of bilateral knee disease, we select 1010 patients with ACLD on the left knee and 1126 patients on the right. Inspired by \cite{xu2018amodel}, a similar treatment of multi-frame dynamic pressure maps is performed, and we obtain the maximum, sum, and effective average of plantar pressures. When given a raw dynamic pressure data $D_{raw}$ containing $K$ frames, the maximum pressure of coordinate $(i,j)$ is $\max_{k\in [1,K]} D_{raw}^k (i,j)$ and the sum one is $\sum_{k=1}^K D_{raw}^k (i,j)$. The formula for calculating the effective average pressure is $\frac{1}{C_{i,j}}\sum_{k=1}^K D_{raw}^k (i,j)$, where $C_{i,j}$ is the total number of frames for which the pressure value at point $(i,j)$ is not zero. We normalize 6 images from one case to the size of $52\times 32$.

Using the method in \cite{xu2018amodel}, we train a deep convolutional neural network that differentiates plantar pressure of healthy and ACLD patients. Then utilize this model to check whether the pressure of N2RPP reconstruction is correct. The network structure is similar to that in \cite{xu2018amodel}, except that the Adam optimizer is used for training. 70\% of the data are used for training, and 10\% are used for cross validation, and the remaining 20\% of the data are used for testing. When training N2RPP, we set the weight parameter $\alpha = 0.03$. The number of training iterations is 20,000 times, and the iteration $k_D$ step of $D$ is 1. We entered all diseased left and right feet rebuilt by N2RPP into the test network, and the results are shown in Tab~\ref{tab:results}. As we can see, the corrected plantar pressure does not differ by more than 8\% from the ground truth with the labels. This indicates that plantar pressure rebuilt by N2RPP approaches the plantar pressure distribution of healthy people to some extent. This result can guide the production of some physical devices, such as insole, to help adjust the plantar pressure distribution of patients, which in turn helps patients recover from surgery or relieve the impact of pain.

% 表内数据待补充。。。
\begin{table}[htb]
\centering
\caption{Results of evaluating N2RPP reconstructed pressure on the supervised model.}
\label{tab:results}
\begin{tabular}{c|ccc|ccc}
\hline
 & \multicolumn{3}{c|}{\textit{Left}} & \multicolumn{3}{c}{\textit{Right}} \\ \hline
Model & \textbf{average} & \textbf{max} & \textbf{sum} & \textbf{average} & \textbf{max} & \textbf{sum} \\ \hline
Acc. of test samples & 90.1\% & 88.6\% & 90.1\% & 90.2\% & 90.6\% & 88.5\% \\
AUC & 0.9642 & 0.9639 & 0.9545 & 0.9611 & 0.9658 & 0.9576 \\
Acc. of rebuild  samples & 82.1\% & \textbf{86.7\%} & 86.4\% & 83.9\% & 83.6\% & \textbf{89.0\%} \\ \hline
\end{tabular}
\end{table}

\section{Visualization}
\begin{figure}[tb]
\centering
\includegraphics[width=1.0\textwidth]{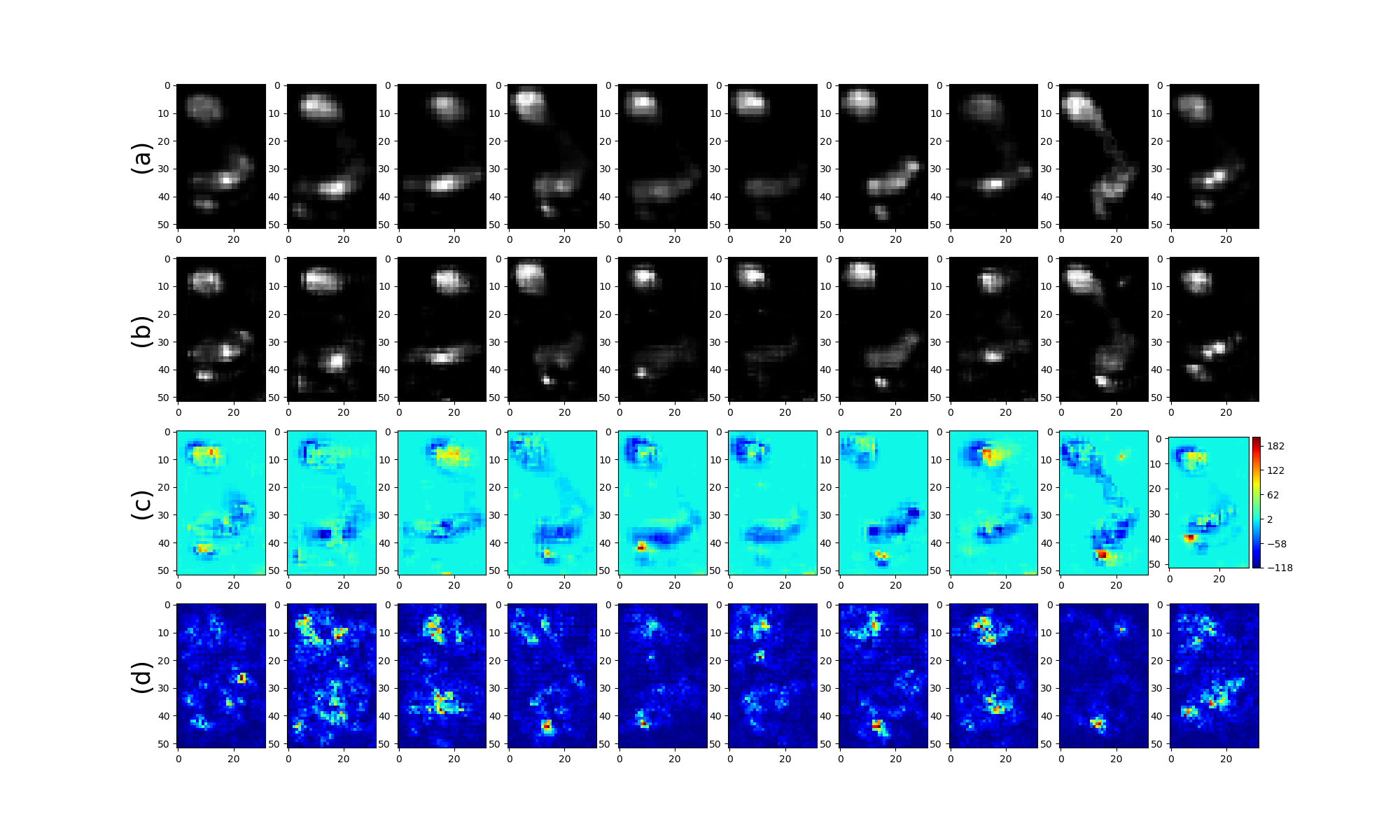}
\caption{Visualization results of N2RPP. (a) are the original plantar pressure images of patients. (b) are the corresponding reconstructed pressure images by N2RPP. (c) are the differences between (a) and (b) calculated by $rebuild\_presure - original\_pressure$. (d) are the saliency maps of (b) for classification drawn by guided backpropagation method.}
\label{fig:vis}
\end{figure}

%少两个图
In this section, we try to discover some details in N2RPP rebuilding the pressure through visualization. We select a model with the highest accuracy and pick out the top 10 feet whose output probability is closest to 1. The visualized results are illustrated in Fig.~\ref{fig:vis}. (b) is the plantar pressure after N2RPP reconstruction. It can be seen that the foot length, foot width, foot progression angle (FPA) and foot shape did not change substantially. The MSE constraint in objective function Eq.~\ref{eq:objective function} plays a certain role. The pressure in some regions has changed, and (c) quantifies these changes and represents them by a heat map. The closer the color is to the red part, the greater the pressure increases and vice versa. It should be noted that this change is only a relative result, because the maximum and minimum pressure values of different samples are not the same. We recorded these two values during preprocessing. Absolute numerical change can be easily restored according to the normalized result. Excluding the impact of noise, we can observe that the increase of pressure often occurs in the toe and heel area. (d) applies guided backpropagation \cite{springenberg2014striving} to visualize the saliency maps for binary classification. We can use these gradients to highlight the areas that have the greatest contribution to the output probability. The bright spots in (d) are the saliency maps that identifies the reconstructed plantar pressure as a positive sample. There are some correspondences between the saliency maps in (d) and the region of changing pressure in (c). This implies that it is the pressure changes at these points that increase the probability of (b) being treated as a healthy sample by the classifier.

\section{Conclusion}
% 写的太烂了。。。
Based on the features extracted from the autoencoder, we propose an adversarial network to rebuild the plantar pressure distribution (N2RPP) for the ACLD patients. The experimental results demonstrate that N2RPP basically learned normal pressure distribution for patients. Through visualization, although some noise appear in generated images, the pressure changes in key regions are obvious. The results may provide some inspirations for making decompression insoles or giving treatment advices by experts.

%\subsubsection*{Acknowledgments.} This paper is partially supported by the National Natural Science Foundation of China (NSFC Grant Nos. 91646202 and 61472006).

\bibliographystyle{splncs}
\bibliography{ref}
%\begin{thebibliography}{4}

%\bibitem{jour} Smith, T.F., Waterman, M.S.: Identification of Common Molecular
%Subsequences. J. Mol. Biol. 147, 195--197 (1981)

%\bibitem{lncschap} May, P., Ehrlich, H.C., Steinke, T.: ZIB Structure Prediction Pipeline:
%Composing a Complex Biological Workflow through Web Services. In: Nagel,
%W.E., Walter, W.V., Lehner, W. (eds.) Euro-Par 2006. LNCS, vol. 4128,
%pp. 1148--1158. Springer, Heidelberg (2006)

%\bibitem{book} Foster, I., Kesselman, C.: The Grid: Blueprint for a New Computing
%Infrastructure. Morgan Kaufmann, San Francisco (1999)

%\bibitem{proceeding1} Czajkowski, K., Fitzgerald, S., Foster, I., Kesselman, C.: Grid
%Information Services for Distributed Resource Sharing. In: 10th IEEE
%International Symposium on High Performance Distributed Computing, pp.
%181--184. IEEE Press, New York (2001)

%\bibitem{proceeding2} Foster, I., Kesselman, C., Nick, J., Tuecke, S.: The Physiology of the
%Grid: an Open Grid Services Architecture for Distributed Systems
%Integration. Technical report, Global Grid Forum (2002)

%\bibitem{url} National Center for Biotechnology Information, \url{http://www.ncbi.nlm.nih.gov}

%\end{thebibliography}

\end{document}